\documentclass{article} 
\usepackage{iclr2023_conference,times}

\usepackage{amsmath,amsfonts,bm}

\def\figref#1{figure~\ref{#1}}

\def\secref#1{section~\ref{#1}}

\def\eqref#1{equation~\ref{#1}}

\def\1{\bm{1}}

\DeclareMathAlphabet{\mathsfit}{\encodingdefault}{\sfdefault}{m}{sl}
\SetMathAlphabet{\mathsfit}{bold}{\encodingdefault}{\sfdefault}{bx}{n}

\usepackage[utf8]{inputenc} 
\usepackage[T1]{fontenc}    
\usepackage{hyperref}       
\usepackage{url}            
\usepackage{booktabs}       
\usepackage{amsfonts}       
\usepackage{nicefrac}       
\usepackage{microtype}      
\usepackage{xcolor}         

\renewcommand{\paragraph}[1]{\vspace{1mm}\noindent\textbf{#1}}
\usepackage{pifont}

\usepackage{adjustbox}
\usepackage{threeparttable}
\usepackage{subfig}
\usepackage{amsmath}
\DeclareCaptionLabelFormat{singleparen}{#2}
\usepackage{multirow}
\usepackage{tabu}
\usepackage{amssymb}

\renewcommand{\figref}[1]{Fig.~\ref{#1}}

\newcommand{\tabref}[1]{Table.~\ref{#1}}

\newcommand{\eqnref}[1]{Eq.~(\ref{#1})}
\renewcommand{\secref}[1]{Sec.~\ref{#1}}
\renewcommand{\paragraph}[1]{\vspace{1mm}\noindent\textbf{#1}}
\newcommand{\ie}{\textit{i.e.}}
\newcommand{\eg}{\textit{e.g.}}
\newcommand{\etal}{\textit{et al}. }

\usepackage{pifont}

\newlength\savewidth\newcommand\shline{\noalign{\global\savewidth\arrayrulewidth
  \global\arrayrulewidth 1pt}\hline\noalign{\global\arrayrulewidth\savewidth}}
\newcommand{\tablestyle}[2]{\setlength{\tabcolsep}{#1}\renewcommand{\arraystretch}{#2}\centering\footnotesize}

\title{Neural Image-based Avatars: Generalizable Radiance Fields for Human Avatar Modeling}

\author{Youngjoong Kwon$^{1}$,\ \ \ Dahun Kim$^{2}$,\ \ \ Duygu Ceylan$^{3}$,\ \ \ Henry Fuchs$^{1}$ \\
$^{1}$University of North Carolina at Chapel Hill. \  $^{2}$Google Research, Brain Team. \  $^{3}$Adobe Research. \\
{\tt \small \{youngjoong,fuchs\}@cs.unc.edu\ \ \{mcahny\}@google.com\ \ \{ceylan\}@adobe.com}
}

\iclrfinalcopy 
\begin{document}

\maketitle

\begin{abstract}
We present a method that enables synthesizing novel views and novel poses of arbitrary human performers from sparse multi-view images. A key ingredient of our method is a hybrid appearance blending module that combines the advantages of the implicit body NeRF representation and image-based rendering. Existing generalizable human NeRF methods that are conditioned on the body model have shown robustness against the geometric variation of arbitrary human performers.
Yet they often exhibit blurry results when generalized onto unseen identities.
Meanwhile, image-based rendering shows high-quality results when sufficient observations are available, whereas it suffers artifacts in sparse-view settings. We propose Neural Image-based Avatars (NIA) that exploits the best of those two methods: to maintain robustness under new articulations and self-occlusions while directly leveraging the available (sparse) source view colors to preserve appearance details of new subject identities.
Our hybrid design outperforms recent methods on both in-domain identity generalization as well as challenging cross-dataset generalization settings. Also, in terms of the pose generalization, our method outperforms even the per-subject optimized animatable NeRF methods. The video results are available at \href{https://youngjoongunc.github.io/nia}{https://youngjoongunc.github.io/nia}.

\end{abstract}

\section{Introduction}

Acquisition of 3D renderable full-body avatars is critical for applications to virtual reality, telepresence and human modeling. While early solutions have required heavy hardware setups such as dense camera rigs or depth sensors, recent neural rendering techniques have achieved significant progress to a more scalable and low-cost solution. Notably, neural radiance fields (NeRF) based methods facilitated by the parametric body prior~\cite{loper2015smpl} require only sparse camera views to enable visually pleasing free-view synthesis~\cite{peng2021neural, kwon2021neural, zhao2022humannerf, cheng2022generalizable} or pose animation~\cite{peng2021animatable, su2021nerf} of the human avatar.

Still, creating a full-body avatar from sparse images (\eg, three snaps) of a person is a challenging problem due to the complexity and diversity of possible human appearances and poses. Most existing methods~\cite{peng2021neural, peng2021animatable} are therefore focusing on person-specific setting which requires a dedicated model optimization for each new subject it encounters. More recent methods explore generalizable human NeRF representations~\cite{peng2021neural,raj2021pva,kwon2021neural,zhao2022humannerf} by using pixel-aligned features in a data-driven manner. Among them, \cite{kwon2021neural} and \cite{chen2022geometry} specifically exploit the \textit{body surface feature} conditioned NeRF (\ie, pixel-aligned features anchored at the SMPL vertices) which helps robustness to various articulations while obviating the need for the 3D supervision~\cite{zhao2022humannerf, cheng2022generalizable}. Nevertheless, these {body surface feature conditioned NeRFs} still suffer blur artifacts when generalizing onto unseen subject identities with complex poses. Also, an extra effort on video-level feature aggregation is required in \cite{kwon2021neural} to compensate for the sparsity of input views.

In this paper, we propose Neural Image-based Avatars (NIA) that generalizes novel view synthesis and pose animation for arbitrary human performers from a sparse set of still images. It is a hybrid framework that combines body surface feature conditioned NeRF (\eg, \cite{kwon2021neural}) and image-based rendering techniques (\eg, \cite{wang2021ibrnet}). While the former helps in robust representation of different body shapes and poses, the image-based rendering helps preserving the color and texture details from the source images. This can complement the NeRF predicted colors which are often blur and inaccurate in generalization settings (cross-identity as well as cross-dataset generalization) as shown in \figref{fig:nvs} and \figref{fig:cross_dataset}. 
To leverage the best of both worlds, we propose a neural appearance blending scheme that learns to adaptively blend the NeRF predicted colors with the direct source image colors. 
Last but not least, by deforming the learned NIA representation based on the skeleton-driven transformations~\cite{lewis2000pose,kavan2007skinning}, we enable plausible pose animation of the learned avatar.

To demonstrate the efficacy of our NIA method, we experiment on ZJU-MoCap~\cite{peng2021neural} and MonoCap~\cite{habermann2020deepcap, habermann2021real} datasets. First, experiments show that our method outperforms the state-of-the-art Neural Human Performer~\cite{kwon2021neural} and GP-NeRF~\cite{chen2022geometry} in novel view synthesis task. Furthermore, we study the more challenging cross-dataset generalization by evaluating the zero-shot performance on the MonoCap~\cite{habermann2020deepcap, habermann2021real} datasets, where we clearly outperform the previous methods. Finally, we evaluate on the pose animation task, where our NIA tested on \textit{unseen subjects} achieves better pose generalization than the \textit{per-subject} trained A-NeRF~\cite{su2021nerf} and Animatable-NeRF~\cite{peng2021animatable} that are tested on the \textit{seen} training subjects.
The ablation studies demonstrate that the proposed modules of our NIA collectively contribute to the high-quality rendering for arbitrary human subjects.

\section{Related Work}
\label{related_work}

Combined with Neural Radiance Fields (NeRF) \cite{mildenhall2020nerf}, human reconstruction research has shown unprecedented development~\cite{pumarola2020d,park2021nerfies,park2021hypernerf}. 
Human priors are utilized to enable robust reconstruction of face and body \cite{Gao-portraitnerf,gafni2021dynamic,peng2021neural}.
However, these methods are per-subject optimized, and cannot model the motions that are not seen during training. Therefore, subsequent works have been focusing on generalization in two directions: pose and subject identity.

\textbf{Pose generalization.} \quad
\cite{su2021nerf} utilize a joint-relative encoding for dynamic articulations. \cite{noguchi2021neural} explicitly associate 3D points to body parts. \cite{chen2021animatable} and \cite{peng2021animatable} deform the target pose space queries into the canonical space to obtain the color and density values. 
\cite{liu2021neural} leverages normal map as the dense pose cue. \cite{xu2021h} learns deformable signed distance field. \cite{weng2022humannerf} and \cite{peng2022animatable} decompose the human deformation into articulation-driven and non-rigid deformations. \cite{su2022danbo} and \cite{zheng2022structured} utilize the joint-specific local radiance fields. 
Despite the significant progress in pose generalization, they still focus on a subject-specific setting which requires training a single model for each subject. In this paper, we tackle generalization across both poses and subject identities. 

\textbf{Subject identity generalization.} \quad
The use of image-conditioned~\cite{rebain2022lolnerf} or pixel-aligned features~\cite{yu2020pixelnerf, wang2021ibrnet} has allowed generalized neural human representations from sparse camera views. \cite{raj2021pva} use camera-encoded pixel-aligned features for face view synthesis. \cite{kwon2021neural} aggregate temporal features by anchoring them onto the SMPL body vertice to complement the sparse input views. \cite{chen2022geometry} also leverages body surface features to enable full-body synthesis. 
\cite{zhao2022humannerf} and \cite{cheng2022generalizable} also propose the neural blending of NeRF prediction with source view colors. Specifically, they use the pixel-aligned features-conditioned NeRF as their implicit body representation. However, pixel-aligned features alone are prone to errors under complex poses as reported in \cite{kwon2021neural}. Therefore, their method require 3D supervision (\eg, depth, visibility) or per-subject finetuning.
In contrast to these methods, we aim at generalizing human modeling without relying on the 3D supervision or per-subject finetuning, but by using only RGB supervision.

\textbf{Other non NeRF-based methods.} \quad Deferred neural rendering based methods \cite{thies2019deferred,raj2021anr,grigorev2021stylepeople} enable fast synthesis of person-specific avatar by combining the traditional graphics pipeline with the neural texture maps. Grigorev \etal further finetunes to deal with unseen subjects. In contrast, we focus on the generalization without any finetuning.
!\cite{kwon2020rotationally,kwon2020mm} exploit volumetric representations to enable free view synthesis from a few snapshots of unseen subjects. 
\cite{saito2019pifu,saito2020pifuhd,he2021arch++} leverage pixel-aligned features to enable 3d reconstruction from a single image, while requiring 3D groundtruth supervision.
\cite{habermann2019livecap,liu2020neural,habermann2021real,bagautdinov2021driving} generate high-quality non-rigid deformation, but they require the template mesh optimization. \cite{aliev2019neural,wu2020multi} anchor features on the point clouds and render them with differentiable rasterizer. However, they require sufficient amount of point clouds that cannot be obtained from our sparse input view setting. In this work, we mainly focus on and compare with the NeRF-based methods.

\begin{figure}[t]
\centering
\def\arraystretch{0.5}
\includegraphics[width=\linewidth]{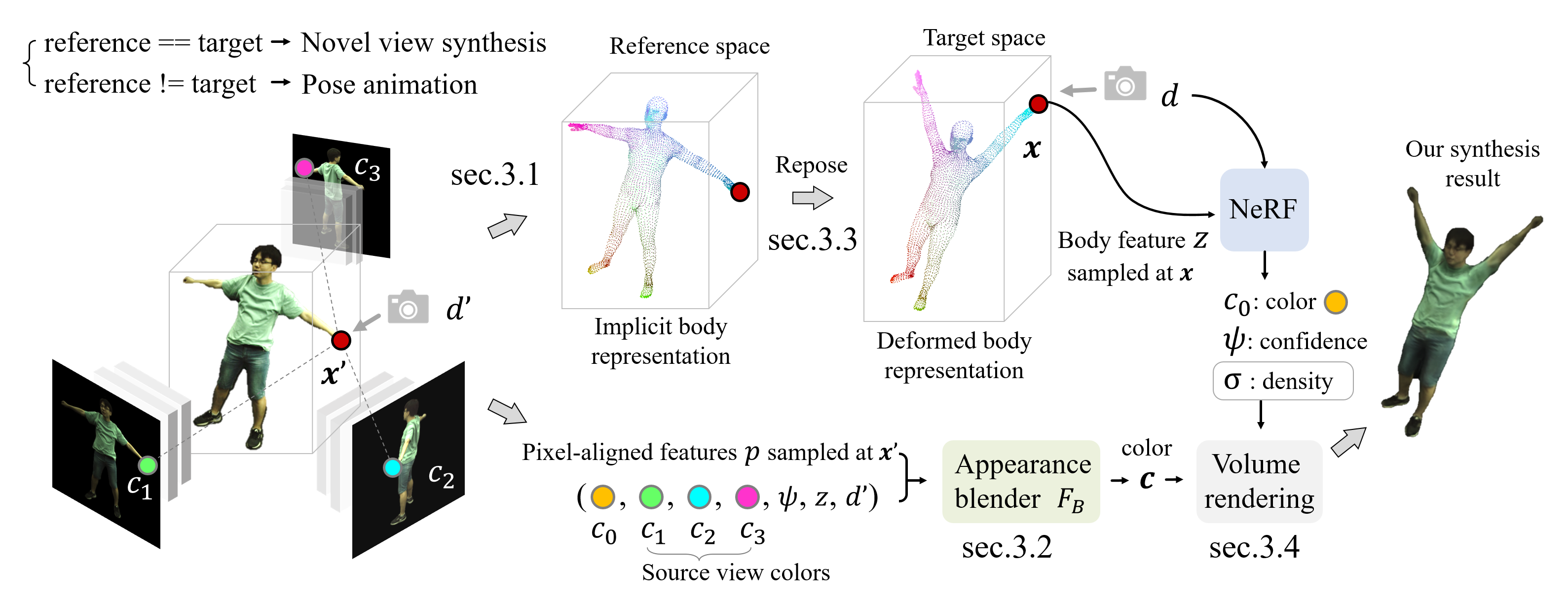}
\caption{\small{\bf Overview of Neural Image-based Avatars (NIA) method.} Given a sparse view images of an unseen person in the reference space, NIA instantly creates a human avatar for novel view synthesis and pose animation. NIA is a hybrid approach combining the SMPL-based implicit body representation and the image-based rendering method. Our appearance blender learns to adaptively blend the predictions from the two components. Note that if the reference space and the target pose spaces are identical (\ie, no pose difference, $x = x'$, $d = d'$), the task is novel view synthesis. Otherwise, it is a pose animation task where we deform the NIA representation to repose the learned avatar.} 
\label{fig:overview}
\vspace{-3mm}
\end{figure}

\section{Method}
\label{method}

The overview of our NIA is given in \figref{fig:overview}. Our goal is to obtain a Neural Image-based Avatars (NIA) which after training, can directly synthesize novel views and novel poses of an unseen person in a feed-forward manner from very sparse input views. To deal with complex and diverse human appearances and poses, we propose a \textit{hybrid} framework that leverages the advantages of two representations. The first is a neural radiance fields based body representation conditioned on a parametric human model (SMPL~\cite{loper2015smpl}), which can provide the body geometry and pose-dependent deformations (\secref{section:implicit_body}). Inspired by \cite{wang2021ibrnet}, the second is obtained by neural image-based rendering which can improve the fine-details of the result by directly fetching the pixel colors from the source view images. A core of our design is a neural appearance blending scheme that learns to blend predictions from the two representations and outputs the final synthesis results (\secref{sec:blending}). Given a set of sparse view images of an arbitrary person, our learned NIA not only performs novel view synthesis, but also reposing by deforming the representation of the avatar at capture time into the desired target pose (\secref{sec:reposing}). 

We note that we assume the calibration parameters of the multi-view input images and the foreground human region masks are known, and the fitted SMPL models are available. 

\subsection{Baseline: Generalizable Implicit Body Representation}
\label{section:implicit_body}
{Inspired by NHP \cite{kwon2021neural} and GP-NeRF \cite{chen2022geometry}, our implicit body NeRF representation follows the design of {body surface feature conditioned NeRF (\ie, pixel-aligned features anchored at SMPL vertices}. 
}

Given $N$ available source views, we extract pixel-aligned image features~\cite{saito2019pifu} from each view $n$ and attach them to the SMPL vertices, to construct a per-view coarse feature set $P_{n}$. Each vertex location is projected to the image feature plane and the corresponding pixel-aligned feature is computed by bilinearly interpolating the grid pixel values. Since $P_{n}$ is sparse in 3D to represent the whole human body, we use 3D sparse convolutions~\cite{liu2015sparse} to diffuse the features into a denser feature volume. The sparse convolutions learn to extrapolate the sparse feature points $P_{n}$ to build a dense body feature volume $\tilde{P}_{n}$ that better adapts to the target body shape. For each query point $\mathbf{x}$, we sample the corresponding body feature from $\tilde{P}_{n}$. The sampled features from $N$ source views are then processed by a view-wise cross-attention with the pixel-aligned features to obtain the multi-view aware body feature $\mathbf{z}_n$. Besides the final density $\sigma$ and color value $\bf{c_0}$, we also propose to predict the confidence $\psi \in [0,1]$ of its color prediction as:
\begin{equation}
\begin{split}
\sigma = F_{\sigma}(\sum_{n}\mathbf{z_n}/N), \quad \quad
{\mathbf{c_{0}},\psi,\mathbf{h}} = F_{\mathbf{c_0}}(\sum_n{(\mathbf{z}_n; \gamma({d}))}/N),
\end{split}
\label{eq:eqn1}
\end{equation}
where $F_{\sigma}$ and $F_{\mathbf{c_0}}$ are MLPs consisting of four linear layers respectively, and $\gamma : \mathbb{R}^{3 \rightarrow {6\times l}}$ is a positional encoding of viewing direction ${d} \in \mathbb{R}^3$ as in \cite{mildenhall2020nerf} with $2 \times l$ different basis functions.
$\mathbf{h}$ is the intermediate color feature extracted from the second to last layer of $F_{\mathbf{c_0}}$ that is later used in the appearance blending (\secref{sec:blending}).

\subsection{Learnable Appearance Blending}
\label{sec:blending}
While our implicit body NeRF representation helps in robustness to geometric variation of arbitrary human shapes, it is difficult for such a general NeRF representation to convey all the high-frequency details (color and textures) present in the source view images. To remedy this, we leverage ideas from image-based rendering techniques~\cite{chen1993view,buehler2001unstructured,gortler1996lumigraph,levoy1996light,debevec1998efficient,hedman2018deep,wang2021ibrnet}. In particular, a query point $\mathbf{x}$ can be projected to $N$ source views to directly retrieve the color values. We propose an appearance blending module $F_{B}$ that learns to predict blending weights among the $N$ source view colors $\mathbf{c}_n$ and the NeRF generated color $\mathbf{c}_0$. 

The inputs to this network are the pixel-aligned features $p_n$ for $\mathbf{x}$, the relative viewing direction $\hat{d}_n$ which is defined by the difference and dot product between the target viewing direction $d$ and the source viewing direction $d_n$, the visibility $o_{n}$ of $\mathbf{x}$ with respect to each source view, the intermediate NeRF color feature $\mathbf{h}$ and the color confidence $\psi$ from $F_{c_{0}}$. We define the view index for the implicit body NeRF representation as $n$=$0$. Formally, the blending weights ${w^{rgb}_n}$ are computed as:
\begin{equation} \label{eq:blend_mlp}
    \{w^{rgb}_n\}_{n=0}^{N}=F_B(\{{p}_{n};{\hat{d}_n};{o_n}\}_{n=1}^{N},\mathbf{h},\psi).
\end{equation}

Note that our visibility $o_n$ is computed \textit{without} the need for any ground truth 3D geometry. To pursue our scalable and practical setting without the reliance on the 3D supervision, we propose to exploit the underlying parametric body model to approximate the visibility for each query point. Specifically, we borrow the vertex visibility of the SMPL model. As in \cite{huang2020arch,bhatnagar2020loopreg,peng2021animatable}, we first search the SMPL mesh surface that is closest to the query point. Then, we define the visibility as a float value $o_n \in [0,1]$ computed by the barycentric interpolation of the 0-or-1 visibility values of three vertices on the corresponding mesh facet. This is different from \cite{cheng2022generalizable}, where visibility trained with groundtruth is required. Also, our design differs from \cite{zhao2022humannerf}, where the groundtruth-supervised depth is utilized to sample source view colors.
The final color $\mathbf{c}$ for $\mathbf{x}$ is computed via softmax as ${\mathbf{c} = \sum_{n=0}^N (\text{exp}(w^{rgb}_n)\mathbf{c}_n)/(\sum_{i=0}^N \text{exp}(w^{rgb}_i))}$.

\subsection{Re-posing NIA for Pose Animation}
\label{sec:reposing}
We describe how our proposed NIA can be extended to animate an unseen subject given new target poses at inference without any further training. Different from the previous works \cite{chen2021animatable,peng2021animatable,peng2022animatable}, where a canonical feature space for appearance retrieval is optimized in a per-subject manner, our generalization approach aims at reposing the NIA avatars that are created on the fly given a set of sparse reference images of unseen human subjects. 
Since the available source images in the reference space cannot be directly used in the target pose space (\ie, observation space), the construction of our NIA representations should be considered with the deformation between the target and reference spaces.

To build the implicit body feature volume in the target pose space, we retrieve the relevant pixel-aligned features from the location of the reference SMPL vertices. By using the known SMPL correspondences, these pixel-aligned feature points are relocated and attached to the corresponding target pose SMPL vertices. The following sparse convolutions will result in the feature volume in the target pose space, where each body feature ${z}^{tar}_n$ is sampled from. The density and color in the target space are computed similarly to \eqnref{eq:eqn1}, except we use ${z}^{tar}_n$ that is transformed into the reference space, instead of $z_n$. 

The appearance blending is defined similarly to \eqnref{eq:blend_mlp} using the target-to-reference transformed query location $\mathbf{x'}$ and viewing direction $d'$, instead of $\mathbf{x}$ and $d$. The visibility is computed with respect to $\mathbf{x'}$ and the reference SMPL body.

\textbf{Deformation modeling.} \quad
We model the deformation $T$ from the target pose space (observation space) to reference space as the skeleton-driven deformation, which is the composition of the inverse and forward linear-blend skinning~\cite{lewis2000pose}. Specifically, we can transform the query $\mathbf{x}$ in the target space into the reference space location $\mathbf{x}'$ through $T(\mathbf{x})=(T^{fwd} \circ T^{inv})(\mathbf{x})$, where $T^{inv}$ and $T^{fwd}$ are defined as:
\begin{equation} \label{eq:transform}
\begin{split}
     T^{inv} = \left( \sum_{k=1}^K w_k^{tar}(\mathbf{x}) G_k \right)^{-1} \mathbf{x} = \mathbf{x}^{can}, \quad \quad
     T^{fwd} = \left( \sum_{k=1}^K w_k^{can}(\mathbf{x}^{can}) G_k \right)^{-1} \mathbf{x}^{can} = \mathbf{x}^{'},     
\end{split}
\end{equation}
where $\{G_k\} \in SE(3)$ is the $K$ transformation matrices produced by SMPL skeleton with $K$ body parts, and $w_k^{tar}$ and $w_k^{can}$ are the blending weights for $k$-th part sampled in the target and canonical space, respectively. Similar to the visibility computation (\secref{sec:blending}), the blending weight is interpolated from the blending weights of the nearest SMPL vertices.

The deformation $\tilde{T}$ for the viewing direction, which transforms the target-space viewing direction $\mathbf{d}$ to the reference space viewing direction $\mathbf{d}'$ is defined in a similar manner. We denote the weighted sum of transformation matrices of $T$ as $[R(\mathbf{x}); t(\mathbf{x})]$. Then, $\tilde{T}(\mathbf{d})$ is defined as $\tilde{T}(\mathbf{d}) = R(\mathbf{x})\mathbf{d} = \mathbf{d}'$.

\subsection{Volume Rendering and Loss Function}
\label{section:render}
\textbf{Volume rendering.}\quad To decide the color of each ray, we accumulate the density and color predictions along the ray $\mathbf{r}(t) =\mathbf{r}_0 + t\mathbf{d}$ for $t \in [t_{\mathrm{near}},t_{\mathrm{far}}]$ as defined in NeRF~\cite{mildenhall2020nerf} as follows:
\begin{equation}
    \mathbf{C}(r) = \int_{t_{\mathrm{near}}}^{t_{\mathrm{far}}} \mathbf{T}(t)\sigma(\mathbf{{r}}(t))\mathbf{c(r}(t),\mathbf{d}) dt, \quad where\quad 
    \mathbf{T}(t) = \operatorname{exp}\left(-\int_{t_{\mathrm{near}}}^{t} \mathbf{\sigma(r}(s))ds\right)
\end{equation}

In practice, we uniformly sample a set of $64$ points $t\sim[t_{near}, t_{far}]$. The bounds for ray sampling $t_{near}, t_{far}$ are drived by computing the 3D bounding box of the SMPL body.

\textbf{Loss function.}\quad
Together with the final blended appearance $\mathbf{c}$, we also supervise the NeRF regressed $\mathbf{c_0}$ with a simple photometric loss to improve the prediction quality: 
$\mathcal{L} = \lVert\mathbf{c}_0 - \mathbf{\hat{c}} \rVert_2 + \lVert\mathbf{c} - \mathbf{\hat{c}} \rVert_2 $ where $\mathbf{\hat{c}}$ denotes the groundtruth pixel color.

\section{Experimental Results}

To demonstrate the effectiveness of our proposed NIA method, we conduct experiments for novel view synthesis and pose animation tasks. We use ZJU-MoCap~\cite{peng2021neural} for both tasks and ablation studies. Then we study our cross-dataset generalization ability by training on ZJU-Mocap and testing on MonoCap datasets without any finetuning.  

\subsection{Identity Generalization}

\paragraph{Competing baselines.}\quad
For novel view synthesis, the state-of-the-art competitors are NHP~\cite{kwon2021neural} and GP-NeRF~\cite{chen2022geometry} which robustly perform generalization with the body surface feature.
Other generalizable methods are pixel-NeRF~\cite{yu2020pixelnerf}, PVA~\cite{raj2021pva}, Image-Based Rendering Network (IBRNet)~\cite{wang2021ibrnet}, and Keypoint NeRF~\cite{mihajlovic2022keypointnerf}. We also compare with per-subject optimization method Neural Body~\cite{peng2021neural}, Neural Textures (NT) \cite{thies2019deferred}, and Neural Human Rendering (NHR) \cite{wu2020multi}.

For pose animation task, we compare with the per-subject animatable NeRF methods which are Neural Body~\cite{peng2021neural}, A-NeRF~\cite{su2021nerf} and Animatable-NeRF~\cite{peng2021animatable}.

Note that the comparison with per-subject methods in both tasks is for reference purpose and places our (and other) generalizable method in disadvantage. These methods require one model (network) trained for a single subject, \ie, as many models as the number of testing subjects, and they are evaluated on the training seen subjects with unseen poses. On the other hand, the generalizable methods train only one model which is evaluated on all unseen subjects with unseen poses.

\begin{figure}[t]
\centering
\includegraphics[width=1.0\linewidth]{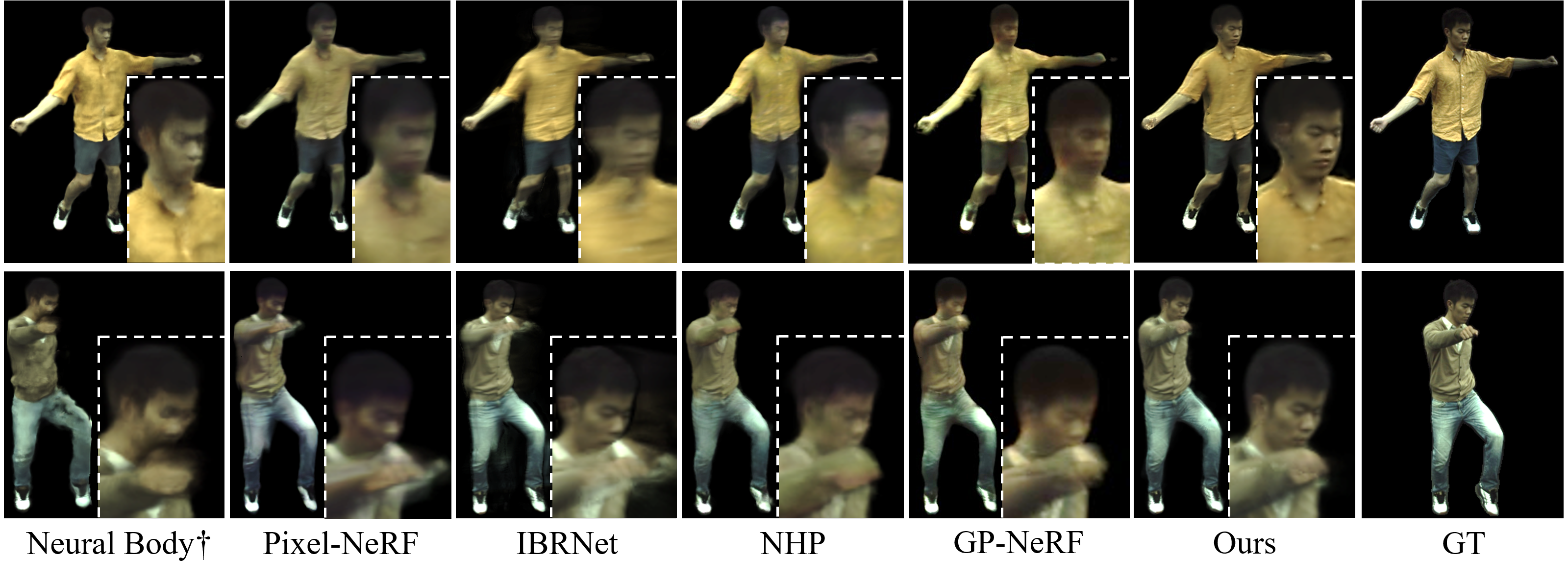}
\vspace{-3mm}
\caption{\small{Novel view synthesis results on ZJU-MoCap dataset. $\dagger$: Neural Body~\cite{peng2021neural} is a per-subject optimized method, and is tested on the seen subjects with unseen poses. Pixel-NeRF~\cite{yu2020pixelnerf}, IBRNet~\cite{wang2021ibrnet}, NHP~\cite{kwon2021neural} and GP-NeRF~\cite{chen2022geometry} are generalizable methods which are tested on unseen subjects with unseen poses.}}
\label{fig:nvs}

\vspace{4mm}
\includegraphics[width=1.0\linewidth]{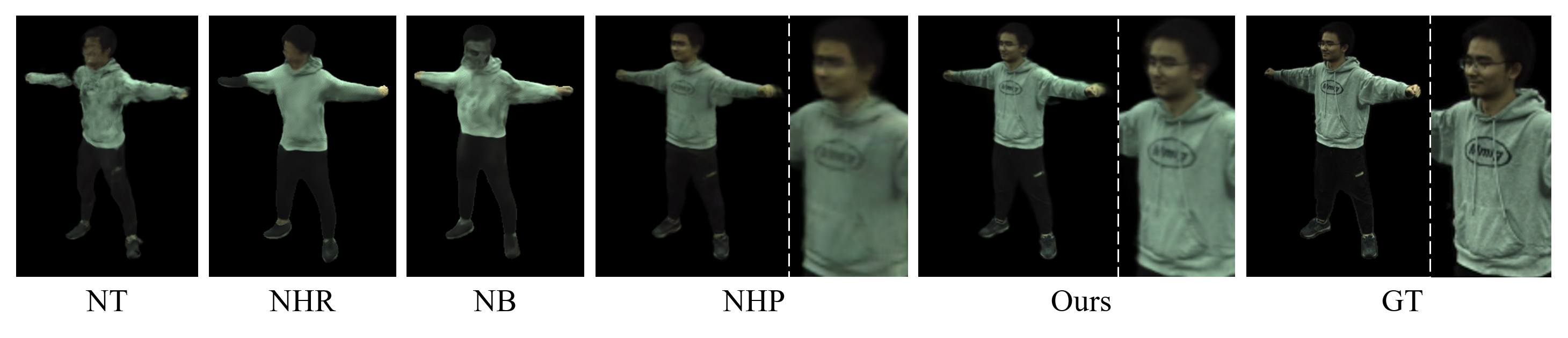}
\vspace{-4mm}
\caption{\small{Results on the unseen poses of seen subjects on ZJU-MoCap dataset. NT: Neural Texture~\cite{thies2019deferred}, NHR: Multi-view Neural Human Rendering~\cite{wu2020multi}, NB: Neural Body~\cite{peng2021neural}, NHP: Neural Human Performer~\cite{kwon2021neural}, and NIA (ours). NT, NHR, NB are per-subject optimized methods.}}
\vspace{-2mm}
\label{fig:persubject}
\end{figure}

\begin{table}
 \centering
  \subfloat[a. Eval. on \textbf{seen} subjects, unseen poses.]{%
  \resizebox{.44\textwidth}{!}{
    \begin{tabular}{llll ccc}
\toprule
{method} &&& PSNR & SSIM & \\ 
\shline
\multicolumn{3}{l}{Training: \textbf{per-subject}} \\ \hline
NT      &&& 22.28 & 0.8720   \\
NHR     &&& 22.31 & 0.8710 & \\
NB      &&& 23.79 & 0.8870 & \\ \hline
\multicolumn{3}{l}{Training: \textbf{generalizable}} \\ \hline
NHP     &&& 26.94 & 0.9290  &\\
\bf{NIA (ours)}   &&& \bf{27.57} & \bf{0.9398} & \\
\bottomrule
\end{tabular}
\label{table:per_subject}
    }

  } \hspace{5mm}
  \subfloat[b. Training: \textbf{generalizable} (\textit{one model for multiple subjects}); Eval. on \textbf{unseen} subjects, unseen poses.]{%
  \resizebox{.47\textwidth}{!}{
    \begin{tabular}{lll|cccc}
\toprule
{method} &&& &PSNR & SSIM \\ 
\shline
PVA              &&&         & 23.15 & 0.8663 & \\
Pixel-NeRF       &      & &  & 23.17 & 0.8693 & \\
IBRNet           &      & &  & 24.54 & 0.8935 & \\
NHP              &      & &  & 24.75 & 0.9058 &\\
GP-NeRF          &      & &  & 24.49 & 0.9012 &\\
Keypoint NeRF    &      & &  & 25.03 & 0.8969 &\\
\bf{NIA (ours)}  &      & &  & \bf{25.79} & \bf{0.9200} & \\
\bottomrule
\end{tabular}
\label{table:generalizable}
    }
  }

\vspace{0mm}
\caption{Novel view synthesis on ZJU-MoCap. All are evaluated on the same testing sequences.}
\label{table:nvs}
\vspace{-2mm}
\end{table}

\begin{figure}
\vspace{4mm}
\includegraphics[width=1.0\linewidth]{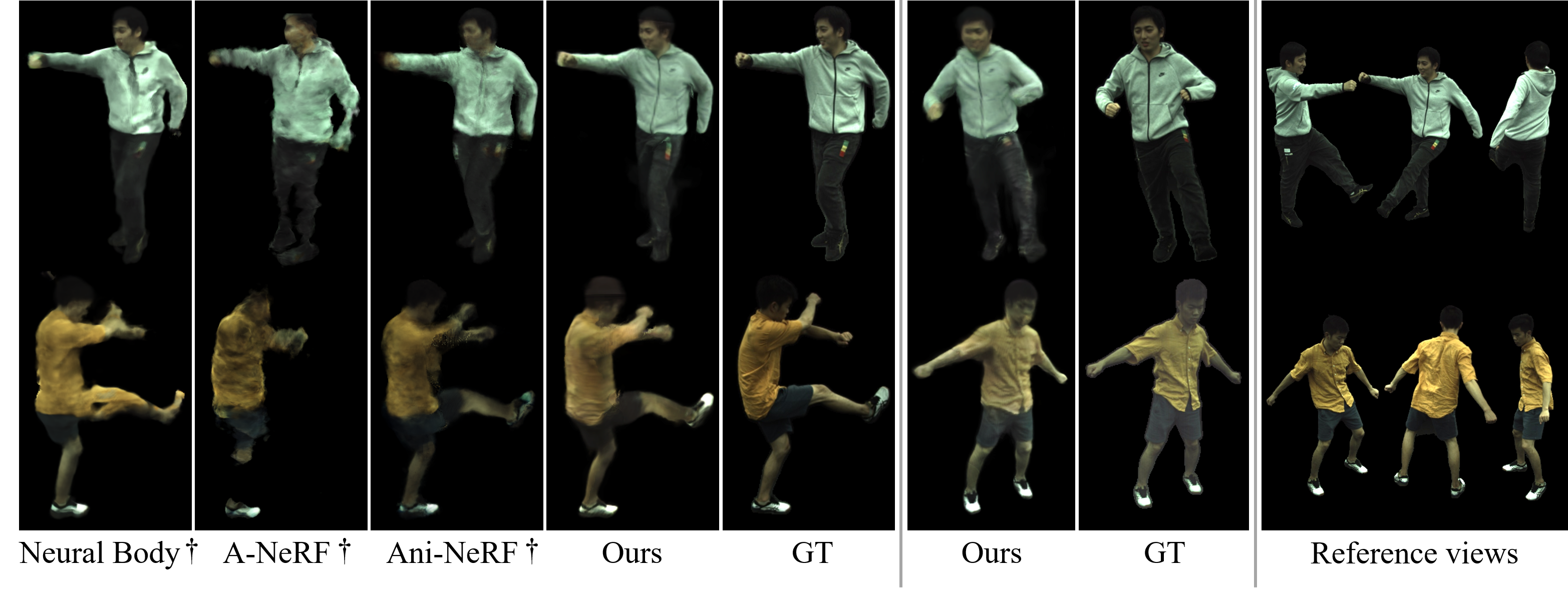}
\caption{Pose animation results on ZJU-MoCap dataset. $\dagger$: Neural Body~\cite{peng2021neural}, A-NeRF~\cite{su2021nerf}, and Ani-NeRF~\cite{peng2021animatable} are per-subject optimized methods, and are evaluated in novel pose synthesis for the seen subjects. Our NIA is the only generalizable method, and is evaluated in novel pose synthesis given only three-view reference images of unseen subjects. Column 6 and 7 are our results for another poses, and the ground truths. The last columns are the reference image input to our NIA.}
\vspace{-1mm}
\label{fig:animation}
\end{figure}

\subsubsection{Novel view synthesis}
\label{sec:nvs}
\paragraph{Setup.}\quad The ZJU-MoCap dataset contains 10 human subjects with 23 synchronized cameras. We follow the same training and testing protocols as in~\cite{kwon2021neural}.

\paragraph{Results.}\quad
The quantitative comparison of our NIA and other methods is shown in \tabref{table:nvs}. Our model achieves state-of-the-art performance on ZJU-Mocap dataset. Especially, the previous state-of-the-art NHP~\cite{kwon2021neural} relies on temporally aggregated features to compensate the sparse input views. Our NIA outperforms NHP by +1.0 dB PSNR and +1.5\% SSIM, while utilizing only still images. NIA also outperforms the most recent state-of-the-art methods GP-NeRF~\cite{chen2022geometry} and Keypoint NeRF~\cite{mihajlovic2022keypointnerf} by +1.3 / +0.7 dB PSNR and +2.0 / +2.5\% SSIM. 

In \tabref{table:nvs} and \figref{fig:persubject}, we show the performance on the unseen poses of the seen subjects. Despite of the disadvantageous generalizable setting (one network for all subjects), our method significantly outperforms other baselines including the per-subject optimized methods.

As shown in \figref{fig:nvs}, our NIA exhibits more high-frequency details compared to NHP and GP-NeRF. Other generalizable methods that utilize image-conditioned features (pixel-NeRF, PVA, and IBRNet) often suffer noisy surface prediction, while NIA shows more robust results on different human shapes.

\subsubsection{Pose animation}
\label{sec:pose_animation}
\paragraph{Setup.}\quad 
We evaluate our NIA for pose animation task on ZJU-MoCap dataset. Given a set of sparse view reference images of an unseen person, our on-the-fly generated avatar for that person is reposed to new target poses. We also use three input views for this task. 

While we do perform generalization onto unseen subject identities, we compare with per-subject methods. Note that this comparison puts our method on \textit{disadvantage} and thus is only to provide a reference level, since the competing methods have \textit{seen} the testing subjects (no identity generalization for baselines).

Specifically, we compare with Neural Body~\cite{peng2021neural}, A-NeRF~\cite{su2021nerf} and Animatable-NeRF~\cite{peng2021animatable} {which made their code public and reported their results on ZJU-MoCap dataset}. Meanwhile, we acknowledge there are more recent and advanced per-subject animation methods~\cite{su2022danbo,zheng2022structured,peng2022animatable,weng2022humannerf}.
For all methods, we evaluate on the same testing sequences as in the novel view synthesis task.

\begin{figure}[t]
\centering
\includegraphics[width=1.0\linewidth]{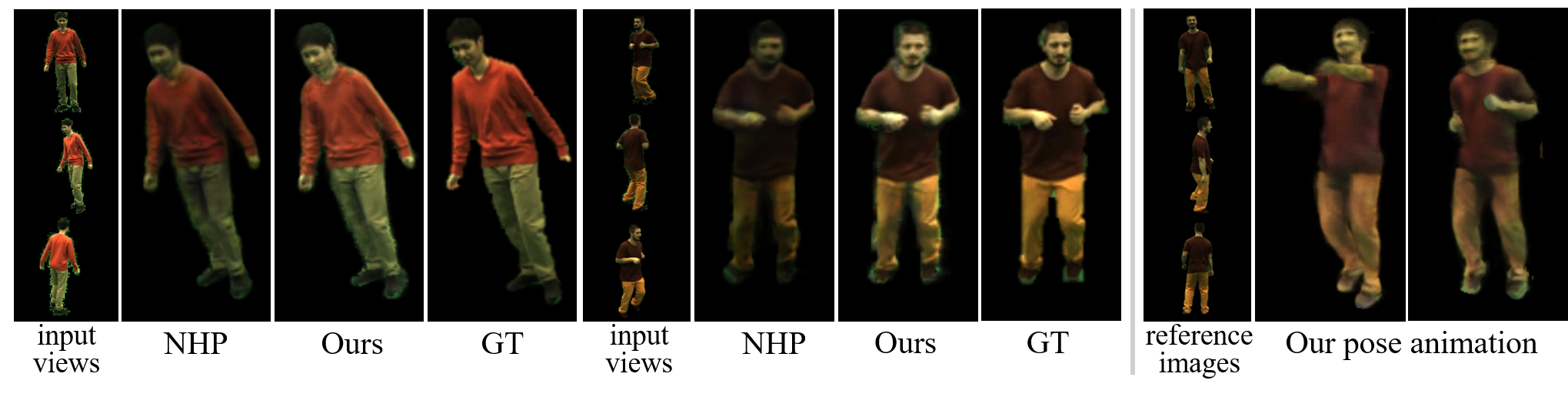}
\vspace{-3mm}
\caption{\small{Cross-dataset generalization results without finetuning. Our NIA outperforms
NHP~\cite{kwon2021neural} on novel view synthesis, when trained on ZJU-MoCap and tested on MonoCap datasets (left). Our NIA can also f
perform pose animation of unseen subjects given only a sparse set of still images (right).}}
\label{fig:cross_dataset}
\vspace{-3mm}
\end{figure}

\begin{table}
\parbox{.68\linewidth}{
\centering
\tablestyle{4pt}{1.0}
\begin{tabular}{l|l|l|cc}
\toprule
method & training & evaluation & PSNR & SSIM \\
\shline
Neural Body       & \multirow{3}{*}{per-subject} & \multirow{3}{*}{\parbox{2.1cm}{\textbf{seen} subjects,\\ {unseen poses}}} & 22.88 & 0.8800 \\
A-NeRF          &  &  & 23.91 & 0.8893 \\
Animatable-NeRF &  &  & 23.88 & 0.8901 \\
\hline
\multirow{2}{*}{\bf{NIA (ours)}} & \multirow{2}{*}{generalizable} & \multirow{2}{*}{\parbox{2.1cm}{\textbf{unseen} subjects,\\ unseen poses}} & \multirow{2}{*}{\bf{24.04}} & \multirow{2}{*}{\bf{0.9000}} 
\\ & & & & \\
\bottomrule
\end{tabular}
\caption{Pose animation on ZJU-MoCap dataset. All methods are evaluated on the same testing sequences and target poses. A-NeRF~\cite{su2021nerf}, Animatable-NeRF~\cite{peng2021animatable}, and Ours.}
\label{tab:animation}
}
\hfill
\parbox{.3\linewidth}{
\centering
\tablestyle{4pt}{1.0}
\begin{tabular}{l|ll}
\toprule
method              & PSNR           & SSIM            \\ \hline
NHP                 & 17.51          & 0.7457          \\
\textbf{NIA (ours)} & \textbf{19.12} & \textbf{0.7804} \\ \hline
\end{tabular}
\caption{\textbf{Cross-dataset} generalization (train on ZJU-MoCap, and test on MonoCap.}
\label{tab:cross_dataset}
}
\vspace{-1mm}
\end{table}

\paragraph{Results.}\quad 
\tabref{tab:animation} shows the quantitative comparison. Remarkably, our NIA model that is generalized onto unseen subjects outperforms the per-subject optimized competitors tested on seen subjects by healthy margins: +1.2 dB / +2.0\% over Neural Body, +0.1 dB / +1.2\% over A-NeRF and +0.1 dB / +1.1\% over Animatable-NeRF in PSNR and SSIM, respectively. In \figref{fig:animation}, we observe that both Neural Body and Animatable-NeRF produce noisy and blurry texture output on novel poses, even though their models are trained to memorize the appearance of the same human subject. Our NIA method preserves relatively more high-frequency details like clothes wrinkles and textures. Note that it is a very challenging setting that requires our model to instantly create animatable avatars on the fly from only three reference view images of unseen human subjects.

\begin{figure}[t]
\centering
\includegraphics[width=\linewidth]{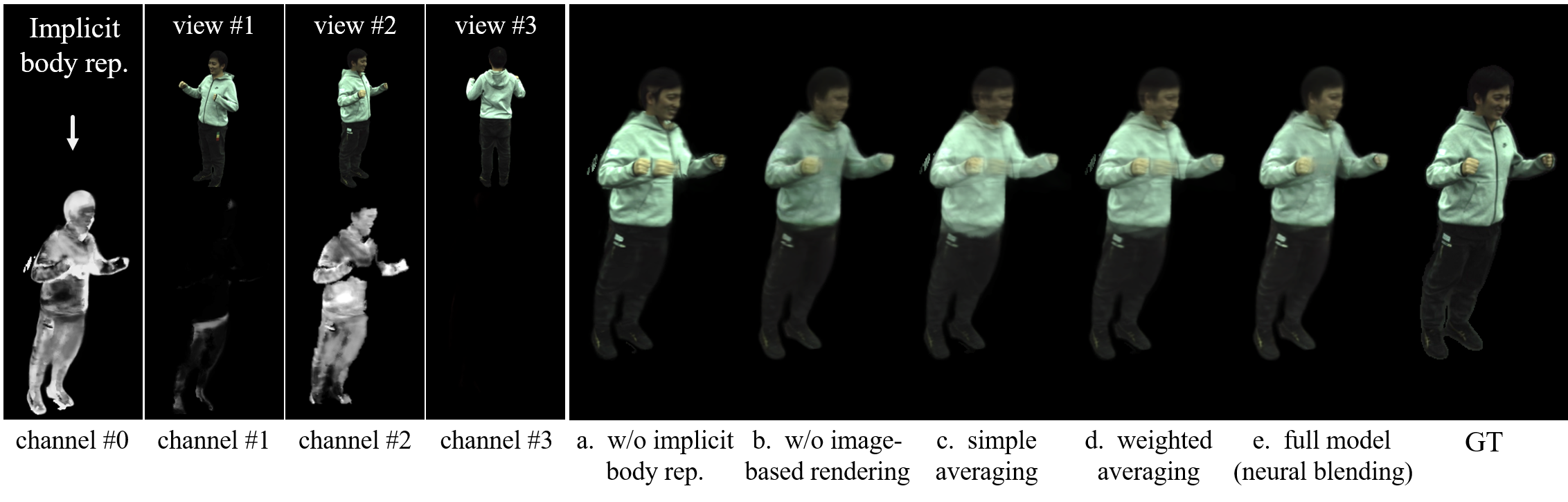}
\caption{\small{Input views (\#1,2,3) and the corresponding channels of learned neural blending weight maps. Channel \#0 corresponds to the prediction from the implicit body representation (NeRF). Remaining columns show the results of ablation items in \tabref{table:ablation2}.i and the ground truth.}}
\vspace{-4mm}
\label{fig:ablation}
\end{figure}

\begin{table}[t]

 \centering
  \tablestyle{4pt}{1.1}
  
  \vspace{2mm}
\subfloat[i. NIA network components.]{%
\begin{tabular}{l|cc}
\toprule
NIA variant & PSNR & SSIM \\
\shline
 
a.\quad w/o implicit body representation
    & {24.62} & {0.8888} \\ 
b.\quad w/o image-based rendering
    & {24.70} & {0.9119} \\
c.\quad w/o neural appearance blending (simple averaging)
    & {25.28} & {0.9168} \\
d.\quad w/o neural appearance blending (weighted averaging)
    & {25.70} & {0.9236} \\   
\hline
e.\quad full model & \bf{26.25} & \bf{0.9268} \\
\bottomrule
\end{tabular}
}
\label{table:ablation1}
  \hspace{3mm}
\subfloat[ii. number of input views.]{%
\begin{tabular}{c|cc}
\toprule
\# views & PSNR & SSIM \\
\shline
1    & {24.71} & {0.9006} \\
2    & {25.53} & {0.9150} \\
3    & {26.25} & {0.9268} \\
\bottomrule
\end{tabular}
}
\caption{Ablation studies on ZJU-MoCap dataset; novel view synthesis task.}
\label{table:ablation2}
\vspace{-3mm}
\end{table}

\subsection{Cross-dataset Generalization}
\label{sec:cross_dataset}
For the out-of-domain, cross-dataset generalization, we train a model on the ZJU-MoCap dataset, and test on the MonoCap dataset without any finetuning. MonoCap consists of DeepCap~\cite{habermann2020deepcap} and DynaCap~\cite{habermann2021real} dataset sequences that are captured by dense camera views. 
Given the different lighting and capture settings between the ZJU-MoCap and MonoCap datasets (\eg, dark v. bright studio), our NIA improves the cross-dataset generalization over NHP by clear margins of +1.6 dB PSNR and +4.5 \% SSIM in \tabref{tab:cross_dataset}.

As shown in the visual results (\figref{fig:cross_dataset}), the implicit body NeRF prediction of NHP especially suffers from the different color distribution between these datasets. 

On the other hand, our NIA is a hybrid model combining the implicit body NeRF with the direct pixel fetching from the input images (\ie image-based rendering), which helps to generalize to out-of-domain MonoCap datasets. {Given only three snaps of a new person, our NIA is able to perform plausible pose animation as shown in \figref{fig:cross_dataset} (right: `pose animation').}

\subsection{Ablation Studies and Analysis}
We provide ablation studies on ZJU-MoCap dataset by evaluating novel view synthesis performance of our NIA variants on unseen subject identities with unseen poses.

\paragraph{NIA architecture designs.}\quad
We study the contribution of different modules of our NIA network in \tabref{table:ablation2}.i and \figref{fig:ablation}. First, we ablate the two main components of our method: implicit body representation (\textit{a}) and image-based rendering (\textit{b}). By removing each component, we observe a performance drop of -1.63 dB / -3.80\% (\textit{a}) and -1.55 dB / -1.49\% (\textit{b}) in PSNR and SSIM respectively.
\figref{fig:ablation} visualizes how each component contributes to the rendering quality. Without the implicit body representation, the model tends to suffer inaccurate shape prediction especially on the regions fully occluded from all source views (\eg, ghosting artifact near the right hand). On the other hand, removing the image-based rendering leads to blurry results and the loss of high-frequency texture details. Our full NIA model-(\textit{e}) leverages these complementary merits and shows the best results both quantitatively and qualitatively.

We further investigate the impact of our neural appearance blending module. It learns to blend predictions from the implicit body representation (\ie  NeRF-predicted color) and the color values from the source images (\ie, image-based rendering), adaptively to different regions to render.
Instead of our learnable design, we experiment by replacing it with two plausible deterministic functions. The first is a naive simple averaging of different color predictions (\textit{c}), and the second is a weighted averaging based on the cosine similarity between each source view's viewing direction and the target view (\textit{d}). The comparison in rows-(\textit{c, d, e}) verifies the benefit of our learnable appearance blender, demonstrated by the gains of +0.97 dB / +1.0\% and +0.45 dB / +0.3\% in PSNR and SSIM over the two deterministic variants respectively. \figref{fig:ablation} also shows the contribution of adaptive blending that helps resolving the blurs and ghosting artifacts.

\paragraph{Number of camera views.}\quad
\tabref{table:ablation2}.ii shows that the performance of our NIA gradually increases with more source views. Increasing the number of source views can provide our model with more relevant observations from the views closer to the target view. 

\textbf{Visualizations.}\quad In \figref{fig:ablation}, we visualize the learned blending weight maps for all source views (\#1, 2, and 3) and the color prediction from the learned implicit body representation (denoted as channel \#0). Among the source views, we observe that the view \#2 which is the most similar to the target view, is highly weighted in large area. Meanwhile, the implicit body representation takes charge of the overall content regions while especially highlighting the occluded regions that are not visible from all source views, (\eg, the left boundary of the torso, and behind the right hand). With our neural appearance blending, our NIA is able to properly blend the available observations and the learned NeRF prediction.

\section{Conclusion}
We introduce Neural Image-based Avatars (NIA), a novel approach for feed-forward avatar creation from sparse view images of unseen persons. Our key idea is to learn to adaptively combine the parametric body model-based NeRF and image-based rendering techniques. Our rendering results in novel view synthesis and pose animation tasks exhibit improved robustness and fine details on different human subjects and poses. Notably, our NIA outperforms the state-of-the-art competitors, and even per-subject optimized methods in both tasks. We believe our approach is a step forward in generalizing the creation of human avatars capable of their free-view synthesis and pose animation.

\section{Acknowledgments}
We thank Sida Peng of Zhejiang University for many very helpful discussions on a variety of implementation details of the Animatable-NeRF. We thank Shih-Yang Su of University of British Columbia for helpful discussions on the A-NeRF details. We thank Prof. Helge Rhodin of UBC and his group for the insightful discussions on the human performance capture. This work was partially supported by National Science Foundation Award 2107454.

\bibliography{iclr2023_conference}

\begin{thebibliography}{59}
\providecommand{\natexlab}[1]{#1}
\providecommand{\url}[1]{\texttt{#1}}
\expandafter\ifx\csname urlstyle\endcsname\relax
  \providecommand{\doi}[1]{doi: #1}\else
  \providecommand{\doi}{doi: \begingroup \urlstyle{rm}\Url}\fi

\bibitem[Aliev et~al.(2019)Aliev, Ulyanov, and Lempitsky]{aliev2019neural}
Kara-Ali Aliev, Dmitry Ulyanov, and Victor Lempitsky.
\newblock Neural point-based graphics.
\newblock \emph{arXiv preprint arXiv:1906.08240}, 2\penalty0 (3):\penalty0 4,
  2019.

\bibitem[Bagautdinov et~al.(2021)Bagautdinov, Wu, Simon, Prada, Shiratori, Wei,
  Xu, Sheikh, and Saragih]{bagautdinov2021driving}
Timur Bagautdinov, Chenglei Wu, Tomas Simon, Fabian Prada, Takaaki Shiratori,
  Shih-En Wei, Weipeng Xu, Yaser Sheikh, and Jason Saragih.
\newblock Driving-signal aware full-body avatars.
\newblock \emph{ACM Transactions on Graphics (TOG)}, 40\penalty0 (4):\penalty0
  1--17, 2021.

\bibitem[Bhatnagar et~al.(2020)Bhatnagar, Sminchisescu, Theobalt, and
  Pons-Moll]{bhatnagar2020loopreg}
Bharat~Lal Bhatnagar, Cristian Sminchisescu, Christian Theobalt, and Gerard
  Pons-Moll.
\newblock Loopreg: Self-supervised learning of implicit surface
  correspondences, pose and shape for 3d human mesh registration.
\newblock \emph{Advances in Neural Information Processing Systems},
  33:\penalty0 12909--12922, 2020.

\bibitem[Buehler et~al.(2001)Buehler, Bosse, McMillan, Gortler, and
  Cohen]{buehler2001unstructured}
Chris Buehler, Michael Bosse, Leonard McMillan, Steven Gortler, and Michael
  Cohen.
\newblock Unstructured lumigraph rendering.
\newblock In \emph{Proceedings of the 28th annual conference on Computer
  graphics and interactive techniques}, pp.\  425--432, 2001.

\bibitem[Chen et~al.(2021)Chen, Zhang, Kang, Zhe, Bao, Jia, and
  Lu]{chen2021animatable}
Jianchuan Chen, Ying Zhang, Di~Kang, Xuefei Zhe, Linchao Bao, Xu~Jia, and
  Huchuan Lu.
\newblock Animatable neural radiance fields from monocular rgb videos, 2021.

\bibitem[Chen et~al.(2022)Chen, Zhang, Xu, Liu, Feng, and
  Yan]{chen2022geometry}
Mingfei Chen, Jianfeng Zhang, Xiangyu Xu, Lijuan Liu, Jiashi Feng, and
  Shuicheng Yan.
\newblock Geometry-guided progressive nerf for generalizable and efficient
  neural human rendering.
\newblock \emph{ECCV}, 2022.

\bibitem[Chen \& Williams(1993)Chen and Williams]{chen1993view}
Shenchang~Eric Chen and Lance Williams.
\newblock View interpolation for image synthesis.
\newblock In \emph{Proceedings of the 20th annual conference on Computer
  graphics and interactive techniques}, pp.\  279--288, 1993.

\bibitem[Cheng et~al.(2022)Cheng, Xu, Piao, Qian, Wu, Lin, and
  Li]{cheng2022generalizable}
Wei Cheng, Su~Xu, Jingtan Piao, Chen Qian, Wayne Wu, Kwan-Yee Lin, and
  Hongsheng Li.
\newblock Generalizable neural performer: Learning robust radiance fields for
  human novel view synthesis.
\newblock \emph{arXiv preprint arXiv:2204.11798}, 2022.

\bibitem[Debevec et~al.(1998)Debevec, Yu, and Borshukov]{debevec1998efficient}
Paul Debevec, Yizhou Yu, and George Borshukov.
\newblock Efficient view-dependent image-based rendering with projective
  texture-mapping.
\newblock In \emph{Eurographics Workshop on Rendering Techniques}, pp.\
  105--116. Springer, 1998.

\bibitem[Dong et~al.(2020)Dong, Shuai, Zhang, Liu, Zhou, and
  Bao]{dong2020motion}
Junting Dong, Qing Shuai, Yuanqing Zhang, Xian Liu, Xiaowei Zhou, and Hujun
  Bao.
\newblock Motion capture from internet videos.
\newblock In \emph{European Conference on Computer Vision}, pp.\  210--227.
  Springer, 2020.

\bibitem[Dong et~al.(2021)Dong, Fang, Jiang, Yang, Bao, and Zhou]{dong2021fast}
Junting Dong, Qi~Fang, Wen Jiang, Yurou Yang, Hujun Bao, and Xiaowei Zhou.
\newblock Fast and robust multi-person 3d pose estimation and tracking from
  multiple views.
\newblock In \emph{T-PAMI}, 2021.

\bibitem[Fang et~al.(2021)Fang, Shuai, Dong, Bao, and Zhou]{fang2021mirrored}
Qi~Fang, Qing Shuai, Junting Dong, Hujun Bao, and Xiaowei Zhou.
\newblock Reconstructing 3d human pose by watching humans in the mirror.
\newblock In \emph{CVPR}, 2021.

\bibitem[Gafni et~al.(2021)Gafni, Thies, Zollhofer, and
  Nie{\ss}ner]{gafni2021dynamic}
Guy Gafni, Justus Thies, Michael Zollhofer, and Matthias Nie{\ss}ner.
\newblock Dynamic neural radiance fields for monocular 4d facial avatar
  reconstruction.
\newblock In \emph{Proceedings of the IEEE/CVF Conference on Computer Vision
  and Pattern Recognition}, pp.\  8649--8658, 2021.

\bibitem[Gao et~al.(2020)Gao, Shih, Lai, Liang, and Huang]{Gao-portraitnerf}
Chen Gao, Yichang Shih, Wei-Sheng Lai, Chia-Kai Liang, and Jia-Bin Huang.
\newblock Portrait neural radiance fields from a single image.
\newblock \emph{arXiv preprint arXiv:2012.05903}, 2020.

\bibitem[Gong et~al.(2018)Gong, Liang, Li, Chen, Yang, and
  Lin]{gong2018instance}
Ke~Gong, Xiaodan Liang, Yicheng Li, Yimin Chen, Ming Yang, and Liang Lin.
\newblock Instance-level human parsing via part grouping network.
\newblock In \emph{Proceedings of the European Conference on Computer Vision
  (ECCV)}, pp.\  770--785, 2018.

\bibitem[Gortler et~al.(1996)Gortler, Grzeszczuk, Szeliski, and
  Cohen]{gortler1996lumigraph}
Steven~J Gortler, Radek Grzeszczuk, Richard Szeliski, and Michael~F Cohen.
\newblock The lumigraph.
\newblock In \emph{Proceedings of the 23rd annual conference on Computer
  graphics and interactive techniques}, pp.\  43--54, 1996.

\bibitem[Grigorev et~al.(2021)Grigorev, Iskakov, Ianina, Bashirov, Zakharkin,
  Vakhitov, and Lempitsky]{grigorev2021stylepeople}
Artur Grigorev, Karim Iskakov, Anastasia Ianina, Renat Bashirov, Ilya
  Zakharkin, Alexander Vakhitov, and Victor Lempitsky.
\newblock Stylepeople: A generative model of fullbody human avatars.
\newblock In \emph{Proceedings of the IEEE/CVF Conference on Computer Vision
  and Pattern Recognition}, pp.\  5151--5160, 2021.

\bibitem[Habermann et~al.(2019)Habermann, Xu, Zollhoefer, Pons-Moll, and
  Theobalt]{habermann2019livecap}
Marc Habermann, Weipeng Xu, Michael Zollhoefer, Gerard Pons-Moll, and Christian
  Theobalt.
\newblock Livecap: Real-time human performance capture from monocular video.
\newblock \emph{ACM Transactions On Graphics (TOG)}, 38\penalty0 (2):\penalty0
  1--17, 2019.

\bibitem[Habermann et~al.(2020)Habermann, Xu, Zollhofer, Pons-Moll, and
  Theobalt]{habermann2020deepcap}
Marc Habermann, Weipeng Xu, Michael Zollhofer, Gerard Pons-Moll, and Christian
  Theobalt.
\newblock Deepcap: Monocular human performance capture using weak supervision.
\newblock In \emph{Proceedings of the IEEE/CVF Conference on Computer Vision
  and Pattern Recognition}, pp.\  5052--5063, 2020.

\bibitem[Habermann et~al.(2021)Habermann, Liu, Xu, Zollhoefer, Pons-Moll, and
  Theobalt]{habermann2021real}
Marc Habermann, Lingjie Liu, Weipeng Xu, Michael Zollhoefer, Gerard Pons-Moll,
  and Christian Theobalt.
\newblock Real-time deep dynamic characters.
\newblock \emph{ACM Transactions on Graphics (TOG)}, 40\penalty0 (4):\penalty0
  1--16, 2021.

\bibitem[He et~al.(2021)He, Xu, Saito, Soatto, and Tung]{he2021arch++}
Tong He, Yuanlu Xu, Shunsuke Saito, Stefano Soatto, and Tony Tung.
\newblock Arch++: Animation-ready clothed human reconstruction revisited.
\newblock In \emph{Proceedings of the IEEE/CVF International Conference on
  Computer Vision}, pp.\  11046--11056, 2021.

\bibitem[Hedman et~al.(2018)Hedman, Philip, Price, Frahm, Drettakis, and
  Brostow]{hedman2018deep}
Peter Hedman, Julien Philip, True Price, Jan-Michael Frahm, George Drettakis,
  and Gabriel Brostow.
\newblock Deep blending for free-viewpoint image-based rendering.
\newblock \emph{ACM Transactions on Graphics (TOG)}, 37\penalty0 (6):\penalty0
  1--15, 2018.

\bibitem[Huang et~al.(2020)Huang, Xu, Lassner, Li, and Tung]{huang2020arch}
Zeng Huang, Yuanlu Xu, Christoph Lassner, Hao Li, and Tony Tung.
\newblock Arch: Animatable reconstruction of clothed humans.
\newblock In \emph{Proceedings of the IEEE/CVF Conference on Computer Vision
  and Pattern Recognition}, pp.\  3093--3102, 2020.

\bibitem[Kavan et~al.(2007)Kavan, Collins, {\v{Z}}{\'a}ra, and
  O'Sullivan]{kavan2007skinning}
Ladislav Kavan, Steven Collins, Ji{\v{r}}{\'\i} {\v{Z}}{\'a}ra, and Carol
  O'Sullivan.
\newblock Skinning with dual quaternions.
\newblock In \emph{Proceedings of the 2007 symposium on Interactive 3D graphics
  and games}, pp.\  39--46, 2007.

\bibitem[Kwon et~al.(2020{\natexlab{a}})Kwon, Petrangeli, Kim, Wang, Fuchs, and
  Swaminathan]{kwon2020mm}
Youngjoong Kwon, Stefano Petrangeli, Dahun Kim, Haoliang Wang, Henry Fuchs, and
  Viswanathan Swaminathan.
\newblock Rotationally-consistent novel view synthesis for humans.
\newblock In \emph{Proceedings of the 28th ACM International Conference on
  Multimedia}, pp.\  2308--2316, 2020{\natexlab{a}}.

\bibitem[Kwon et~al.(2020{\natexlab{b}})Kwon, Petrangeli, Kim, Wang, Park,
  Swaminathan, and Fuchs]{kwon2020rotationally}
Youngjoong Kwon, Stefano Petrangeli, Dahun Kim, Haoliang Wang, Eunbyung Park,
  Viswanathan Swaminathan, and Henry Fuchs.
\newblock Rotationally-temporally consistent novel view synthesis of human
  performance video.
\newblock In \emph{European Conference on Computer Vision}, pp.\  387--402.
  Springer, 2020{\natexlab{b}}.

\bibitem[Kwon et~al.(2021)Kwon, Kim, Ceylan, and Fuchs]{kwon2021neural}
Youngjoong Kwon, Dahun Kim, Duygu Ceylan, and Henry Fuchs.
\newblock Neural human performer: Learning generalizable radiance fields for
  human performance rendering.
\newblock \emph{Advances in Neural Information Processing Systems}, 34, 2021.

\bibitem[Levoy \& Hanrahan(1996)Levoy and Hanrahan]{levoy1996light}
Marc Levoy and Pat Hanrahan.
\newblock Light field rendering.
\newblock In \emph{Proceedings of the 23rd annual conference on Computer
  graphics and interactive techniques}, pp.\  31--42, 1996.

\bibitem[Lewis et~al.(2000)Lewis, Cordner, and Fong]{lewis2000pose}
John~P Lewis, Matt Cordner, and Nickson Fong.
\newblock Pose space deformation: a unified approach to shape interpolation and
  skeleton-driven deformation.
\newblock In \emph{Proceedings of the 27th annual conference on Computer
  graphics and interactive techniques}, pp.\  165--172, 2000.

\bibitem[Liu et~al.(2015)Liu, Wang, Foroosh, Tappen, and Pensky]{liu2015sparse}
Baoyuan Liu, Min Wang, Hassan Foroosh, Marshall Tappen, and Marianna Pensky.
\newblock Sparse convolutional neural networks.
\newblock In \emph{Proceedings of the IEEE conference on computer vision and
  pattern recognition}, pp.\  806--814, 2015.

\bibitem[Liu et~al.(2020)Liu, Xu, Habermann, Zollh{\"o}fer, Bernard, Kim, Wang,
  and Theobalt]{liu2020neural}
Lingjie Liu, Weipeng Xu, Marc Habermann, Michael Zollh{\"o}fer, Florian
  Bernard, Hyeongwoo Kim, Wenping Wang, and Christian Theobalt.
\newblock Neural human video rendering by learning dynamic textures and
  rendering-to-video translation.
\newblock \emph{arXiv preprint arXiv:2001.04947}, 2020.

\bibitem[Liu et~al.(2021)Liu, Habermann, Rudnev, Sarkar, Gu, and
  Theobalt]{liu2021neural}
Lingjie Liu, Marc Habermann, Viktor Rudnev, Kripasindhu Sarkar, Jiatao Gu, and
  Christian Theobalt.
\newblock Neural actor: Neural free-view synthesis of human actors with pose
  control.
\newblock \emph{ACM Transactions on Graphics (TOG)}, 40\penalty0 (6):\penalty0
  1--16, 2021.

\bibitem[Loper et~al.(2015)Loper, Mahmood, Romero, Pons-Moll, and
  Black]{loper2015smpl}
Matthew Loper, Naureen Mahmood, Javier Romero, Gerard Pons-Moll, and Michael~J
  Black.
\newblock Smpl: A skinned multi-person linear model.
\newblock \emph{ACM transactions on graphics (TOG)}, 34\penalty0 (6):\penalty0
  1--16, 2015.

\bibitem[Mihajlovic et~al.(2022)Mihajlovic, Bansal, Zollhoefer, Tang, and
  Saito]{mihajlovic2022keypointnerf}
Marko Mihajlovic, Aayush Bansal, Michael Zollhoefer, Siyu Tang, and Shunsuke
  Saito.
\newblock Keypointnerf: Generalizing image-based volumetric avatars using
  relative spatial encoding of keypoints.
\newblock In \emph{European Conference on Computer Vision (ECCV 2022)}, 2022.

\bibitem[Mildenhall et~al.(2020)Mildenhall, Srinivasan, Tancik, Barron,
  Ramamoorthi, and Ng]{mildenhall2020nerf}
Ben Mildenhall, Pratul~P Srinivasan, Matthew Tancik, Jonathan~T Barron, Ravi
  Ramamoorthi, and Ren Ng.
\newblock Nerf: Representing scenes as neural radiance fields for view
  synthesis.
\newblock In \emph{European Conference on Computer Vision}, pp.\  405--421.
  Springer, 2020.

\bibitem[Noguchi et~al.(2021)Noguchi, Sun, Lin, and Harada]{noguchi2021neural}
Atsuhiro Noguchi, Xiao Sun, Stephen Lin, and Tatsuya Harada.
\newblock Neural articulated radiance field.
\newblock In \emph{Proceedings of the IEEE/CVF International Conference on
  Computer Vision}, pp.\  5762--5772, 2021.

\bibitem[Park et~al.(2021{\natexlab{a}})Park, Sinha, Barron, Bouaziz, Goldman,
  Seitz, and Martin-Brualla]{park2021nerfies}
Keunhong Park, Utkarsh Sinha, Jonathan~T Barron, Sofien Bouaziz, Dan~B Goldman,
  Steven~M Seitz, and Ricardo Martin-Brualla.
\newblock Nerfies: Deformable neural radiance fields.
\newblock In \emph{Proceedings of the IEEE/CVF International Conference on
  Computer Vision}, pp.\  5865--5874, 2021{\natexlab{a}}.

\bibitem[Park et~al.(2021{\natexlab{b}})Park, Sinha, Hedman, Barron, Bouaziz,
  Goldman, Martin-Brualla, and Seitz]{park2021hypernerf}
Keunhong Park, Utkarsh Sinha, Peter Hedman, Jonathan~T Barron, Sofien Bouaziz,
  Dan~B Goldman, Ricardo Martin-Brualla, and Steven~M Seitz.
\newblock Hypernerf: A higher-dimensional representation for topologically
  varying neural radiance fields.
\newblock \emph{arXiv preprint arXiv:2106.13228}, 2021{\natexlab{b}}.

\bibitem[Peng et~al.(2021{\natexlab{a}})Peng, Dong, Wang, Zhang, Shuai, Zhou,
  and Bao]{peng2021animatable}
Sida Peng, Junting Dong, Qianqian Wang, Shangzhan Zhang, Qing Shuai, Xiaowei
  Zhou, and Hujun Bao.
\newblock Animatable neural radiance fields for modeling dynamic human bodies.
\newblock In \emph{ICCV}, 2021{\natexlab{a}}.

\bibitem[Peng et~al.(2021{\natexlab{b}})Peng, Zhang, Xu, Wang, Shuai, Bao, and
  Zhou]{peng2021neural}
Sida Peng, Yuanqing Zhang, Yinghao Xu, Qianqian Wang, Qing Shuai, Hujun Bao,
  and Xiaowei Zhou.
\newblock Neural body: Implicit neural representations with structured latent
  codes for novel view synthesis of dynamic humans.
\newblock In \emph{CVPR}, 2021{\natexlab{b}}.

\bibitem[Peng et~al.(2022)Peng, Zhang, Xu, Geng, Jiang, Bao, and
  Zhou]{peng2022animatable}
Sida Peng, Shangzhan Zhang, Zhen Xu, Chen Geng, Boyi Jiang, Hujun Bao, and
  Xiaowei Zhou.
\newblock Animatable neural implicit surfaces for creating avatars from videos.
\newblock \emph{arXiv preprint arXiv:2203.08133}, 2022.

\bibitem[Pumarola et~al.(2020)Pumarola, Corona, Pons-Moll, and
  Moreno-Noguer]{pumarola2020d}
Albert Pumarola, Enric Corona, Gerard Pons-Moll, and Francesc Moreno-Noguer.
\newblock D-nerf: Neural radiance fields for dynamic scenes.
\newblock \emph{arXiv preprint arXiv:2011.13961}, 2020.

\bibitem[Raj et~al.(2021{\natexlab{a}})Raj, Tanke, Hays, Vo, Stoll, and
  Lassner]{raj2021anr}
Amit Raj, Julian Tanke, James Hays, Minh Vo, Carsten Stoll, and Christoph
  Lassner.
\newblock Anr: Articulated neural rendering for virtual avatars.
\newblock In \emph{Proceedings of the IEEE/CVF Conference on Computer Vision
  and Pattern Recognition}, pp.\  3722--3731, 2021{\natexlab{a}}.

\bibitem[Raj et~al.(2021{\natexlab{b}})Raj, Zollhoefer, Simon, Saragih, Saito,
  Hays, and Lombardi]{raj2021pva}
Amit Raj, Michael Zollhoefer, Tomas Simon, Jason Saragih, Shunsuke Saito, James
  Hays, and Stephen Lombardi.
\newblock Pva: Pixel-aligned volumetric avatars.
\newblock \emph{arXiv preprint arXiv:2101.02697}, 2021{\natexlab{b}}.

\bibitem[Rebain et~al.(2022)Rebain, Matthews, Yi, Lagun, and
  Tagliasacchi]{rebain2022lolnerf}
Daniel Rebain, Mark Matthews, Kwang~Moo Yi, Dmitry Lagun, and Andrea
  Tagliasacchi.
\newblock Lolnerf: Learn from one look.
\newblock In \emph{Proceedings of the IEEE/CVF Conference on Computer Vision
  and Pattern Recognition}, pp.\  1558--1567, 2022.

\bibitem[Saito et~al.(2019)Saito, Huang, Natsume, Morishima, Kanazawa, and
  Li]{saito2019pifu}
Shunsuke Saito, Zeng Huang, Ryota Natsume, Shigeo Morishima, Angjoo Kanazawa,
  and Hao Li.
\newblock Pifu: Pixel-aligned implicit function for high-resolution clothed
  human digitization.
\newblock In \emph{Proceedings of the IEEE/CVF International Conference on
  Computer Vision}, pp.\  2304--2314, 2019.

\bibitem[Saito et~al.(2020)Saito, Simon, Saragih, and Joo]{saito2020pifuhd}
Shunsuke Saito, Tomas Simon, Jason Saragih, and Hanbyul Joo.
\newblock Pifuhd: Multi-level pixel-aligned implicit function for
  high-resolution 3d human digitization.
\newblock In \emph{Proceedings of the IEEE/CVF Conference on Computer Vision
  and Pattern Recognition}, pp.\  84--93, 2020.

\bibitem[Schwarz et~al.(2020)Schwarz, Liao, Niemeyer, and
  Geiger]{Schwarz2020NEURIPS}
Katja Schwarz, Yiyi Liao, Michael Niemeyer, and Andreas Geiger.
\newblock Graf: Generative radiance fields for 3d-aware image synthesis.
\newblock In \emph{Advances in Neural Information Processing Systems
  (NeurIPS)}, 2020.

\bibitem[Su et~al.(2021)Su, Yu, Zollh{\"o}fer, and Rhodin]{su2021nerf}
Shih-Yang Su, Frank Yu, Michael Zollh{\"o}fer, and Helge Rhodin.
\newblock A-nerf: Articulated neural radiance fields for learning human shape,
  appearance, and pose.
\newblock \emph{Advances in Neural Information Processing Systems}, 34, 2021.

\bibitem[Su et~al.(2022)Su, Bagautdinov, and Rhodin]{su2022danbo}
Shih-Yang Su, Timur Bagautdinov, and Helge Rhodin.
\newblock Danbo: Disentangled articulated neural body representations via graph
  neural networks.
\newblock \emph{arXiv preprint arXiv:2205.01666}, 2022.

\bibitem[Thies et~al.(2019)Thies, Zollh{\"o}fer, and
  Nie{\ss}ner]{thies2019deferred}
Justus Thies, Michael Zollh{\"o}fer, and Matthias Nie{\ss}ner.
\newblock Deferred neural rendering: Image synthesis using neural textures.
\newblock \emph{ACM Transactions on Graphics (TOG)}, 38\penalty0 (4):\penalty0
  1--12, 2019.

\bibitem[Vaswani et~al.(2017)Vaswani, Shazeer, Parmar, Uszkoreit, Jones, Gomez,
  Kaiser, and Polosukhin]{vaswani2017attention}
Ashish Vaswani, Noam Shazeer, Niki Parmar, Jakob Uszkoreit, Llion Jones,
  Aidan~N Gomez, Lukasz Kaiser, and Illia Polosukhin.
\newblock Attention is all you need.
\newblock \emph{arXiv preprint arXiv:1706.03762}, 2017.

\bibitem[Wang et~al.(2021)Wang, Wang, Genova, Srinivasan, Zhou, Barron,
  Martin-Brualla, Snavely, and Funkhouser]{wang2021ibrnet}
Qianqian Wang, Zhicheng Wang, Kyle Genova, Pratul Srinivasan, Howard Zhou,
  Jonathan~T Barron, Ricardo Martin-Brualla, Noah Snavely, and Thomas
  Funkhouser.
\newblock Ibrnet: Learning multi-view image-based rendering.
\newblock \emph{arXiv preprint arXiv:2102.13090}, 2021.

\bibitem[Weng et~al.(2022)Weng, Curless, Srinivasan, Barron, and
  Kemelmacher-Shlizerman]{weng2022humannerf}
Chung-Yi Weng, Brian Curless, Pratul~P Srinivasan, Jonathan~T Barron, and Ira
  Kemelmacher-Shlizerman.
\newblock Humannerf: Free-viewpoint rendering of moving people from monocular
  video.
\newblock \emph{arXiv preprint arXiv:2201.04127}, 2022.

\bibitem[Wu et~al.(2020)Wu, Wang, Hu, and Yu]{wu2020multi}
Minye Wu, Yuehao Wang, Qiang Hu, and Jingyi Yu.
\newblock Multi-view neural human rendering.
\newblock In \emph{Proceedings of the IEEE/CVF Conference on Computer Vision
  and Pattern Recognition}, pp.\  1682--1691, 2020.

\bibitem[Xu et~al.(2021)Xu, Alldieck, and Sminchisescu]{xu2021h}
Hongyi Xu, Thiemo Alldieck, and Cristian Sminchisescu.
\newblock H-nerf: Neural radiance fields for rendering and temporal
  reconstruction of humans in motion.
\newblock \emph{Advances in Neural Information Processing Systems}, 34, 2021.

\bibitem[Yu et~al.(2020)Yu, Ye, Tancik, and Kanazawa]{yu2020pixelnerf}
Alex Yu, Vickie Ye, Matthew Tancik, and Angjoo Kanazawa.
\newblock pixelnerf: Neural radiance fields from one or few images.
\newblock \emph{arXiv preprint arXiv:2012.02190}, 2020.

\bibitem[Zhao et~al.(2022)Zhao, Yang, Zhang, Lin, Zhang, Yu, and
  Xu]{zhao2022humannerf}
Fuqiang Zhao, Wei Yang, Jiakai Zhang, Pei Lin, Yingliang Zhang, Jingyi Yu, and
  Lan Xu.
\newblock Humannerf: Efficiently generated human radiance field from sparse
  inputs.
\newblock In \emph{Proceedings of the IEEE/CVF Conference on Computer Vision
  and Pattern Recognition}, pp.\  7743--7753, 2022.

\bibitem[Zheng et~al.(2022)Zheng, Huang, Yu, Zhang, Guo, and
  Liu]{zheng2022structured}
Zerong Zheng, Han Huang, Tao Yu, Hongwen Zhang, Yandong Guo, and Yebin Liu.
\newblock Structured local radiance fields for human avatar modeling.
\newblock \emph{arXiv preprint arXiv:2203.14478}, 2022.

\end{thebibliography}
\bibliographystyle{iclr2023_conference}
\clearpage

\appendix
\section{Video results}

Video results of free-viewpoint rendering, pose animation and 3D reconstruction with ZJU-MoCap and MonoCap datasets can be found at \href{https://youngjoongunc.github.io/nia}{https://youngjoongunc.github.io/nia}. We compare our NIA with IBRNet~\cite{wang2021ibrnet}, NHP~\cite{kwon2021neural}, and Animatable-NeRF~\cite{peng2021animatable}.

\section{Reproducibility}
We describe the implementation details in the interest of reproducibility.

\subsection{Input details.}
Our model requires RGB images, foreground mask, SMPL parameters, and camera calibration parameters. We compute the visibility of all SMPL vertices through rasterization using the calibration and SMPL parameters. We elaborate the used dataset details and input preparation in the Section 3.

\subsection{Implementation details.} 

\paragraph{Image feature extractor.}\quad
Given $N$ source images of shape H$\times$W, we use an ImageNet-pretrained ResNet-18 backbone to extract a feature pyramid. We take the multi-scale feature maps of shapes $\{$ 64 $\times$ H/2 $\times$ W/2, 64 $\times$ H/4 $\times$ W/4, 128 $\times$ H/8 $\times$ W/8, 256 $\times$ H/16 $\times$ W/16 $\}$.
These feature maps are bilinearly upsampled to the highest resolution \ie, H/2 $\times$ W/2, and concatenated into a shape 512 $\times$ H/2 $\times$ W/2.

\paragraph{Construction of the Implicit Body Representation.}\quad
As described in Section 3.1, we first define our body features anchored to the SMPL vertices by attaching the vertices' pixel-aligned features from $N$ source views. Since the body features, which are defined on the SMPL mesh surface, are relatively sparse in the 3D space, we cannot directly interpolate the body features corresponding to query location. Inspired by Peng et al.~\cite{peng2021neural}, we diffuse the body features into a volume and sample the feature at the query location. Specifically, we first compute the bounding box of the SMPL model, where the side length is enlarged by $2.5\%$ to cover the gap between the exact human geometry and the SMPL model, and divide it with small voxel of size $5\times 5\times 5$ $mm^{3}$. This results in the feature volume $P_{n} \in \mathbb{R}^{D\times H\times W}$ (depth, height, width) for each view $n$. Then a series of 3D sparse convolutions~\cite{liu2015sparse} is applied on this volume of sparse feature points (vertices) to densify them within the volume space, resulting in the output feature volume $\tilde{P}_{n}\in \mathbb{R}^{{D\over{16}}\times {H\over{16}}\times {W\over{16}}\times C}$. We use $C = 256$. To make the feature diffusion invariant to the human position or orientation in the world coordinate system, both the body feature and query locations are transformed to the SMPL coordinate system.

Given a query point $\mathbf{x}$, we sample the corresponding body feature $\mathbf{s}_{n} \in \mathbb{R}^{C}$ from the volume $\tilde{P}_{n}$. The sampled body features from $N$ source views $S = \{\mathbf{s}_{n}\}_{n=1}^{N} \in \mathbb{R}^{N\times C}$ are then processed by a view-wise cross-attention with the pixel-aligned features $\mathit{L} = \{\mathbf{\mathit{l}}_{n}\}_{n=1}^{N} \in \mathbb{R}^{N\times C}$ to obtain the multi-view aware body features $Z$ as:
\begin{equation} \label{eq:multiview_attention}
     Z = softmax({1\over{\sqrt{d}}} \phi_{k}(S)\cdot \phi_{k}(\mathit{L})^{T}) \cdot \phi_{v}(\mathit{L}) + S, \quad\quad Z = \{\mathbf{z}_{n}\}_{n=1}^{N} \in \mathbb{R}^{N\times C},
\end{equation}
where $\phi_k$ and $\phi_v$ are the key and value embedding functions as in the Transformers~\cite{vaswani2017attention}.
Besides the final density $\sigma$ and color value $\bf{c_0}$, we also propose to predict the confidence $\psi \in [0,1]$ of its color prediction as:
\begin{equation} \label{eq:nerf}
     \sigma = F_{\sigma}(\sum_{n}\mathbf{z_n}/N), \quad \quad
{\mathbf{c_{0}},\psi,\mathbf{h}} = F_{\mathbf{c_0}}(\sum_n{(\mathbf{z}_n; \gamma({d}))}/N),
\end{equation}
where $F_{\sigma}$ and $F_{\mathbf{c_0}}$ are MLPs consisting of four linear layers respectively, and $\gamma : \mathbb{R}^{3 \rightarrow {6\times l}}$ is a positional encoding of viewing direction ${d} \in \mathbb{R}^3$ as in \cite{mildenhall2020nerf} with $2 \times l$ different basis functions.
$\mathbf{h}$ is the intermediate color feature extracted from the second to last layer of $F_{\mathbf{c_0}}$ that is later used in the appearance blender (Section 3.2).

\paragraph{Query point sampling details.}\quad
We sample the query points inside the 3D bounding box of the SMPL body with its side length enlarged by $2.5\%$ to embrace the gap between real geometry and naked SMPL body. We sample 1024 rays, and 64 points are sampled per ray for the training. For the inference, 64 points are sampled along each ray.

\paragraph{Appearance blending module.}\quad
Our appearance blending module $F_{B}$ is a 8-layer MLP with ReLU activations.

\subsection{Training details.}\quad
Our model is trained for 250 epochs for ZJU-MoCap. The learning rate is set to $5 \times 10^{-4}$. We use one RTX 3090 GPU for training, which takes 1 day.

\section{Datasets} 

In this section, we discuss the additional details about the datasets used, including the train/test splits and their license information. We would like to clarify that both the ZJU-Mocap and MonoCap datasets are public for research purposes and do not contain any personally identifiable information or offensive content.

\subsection{ZJU-MoCap}
ZJU-Mocap\cite{peng2021neural} is the public dataset that is available for the research and non-profit purposes only as stated in their Github page\footnote{https://github.com/zju3dv/neuralbody}. It provides RGB videos of 10 different human subjects captured from 23 synchronized cameras. Each video length is between 600 to 1000 frames. The ZJU-Mocap dataset provides foreground region masks computed by PGN method~\cite{gong2018instance}, SMPL parameters fitted by EasyMocap\footnote{https://github.com/zju3dv/EasyMocap}~\cite{dong2020motion,peng2021neural,fang2021mirrored,dong2021fast}, and camera calibrations. We follow the training and testing protocols of NHP~\cite{kwon2021neural} including the data format and train/test split which are available in their Github page\footnote{https://github.com/YoungJoongUNC/Neural\_Human\_Performer}.

\subsection{MonoCap}
MonoCap is the public dataset that is available for the research and non-profit purposes only as stated in their website\footnote{https://gvv-assets.mpi-inf.mpg.de}. MonoCap consists of DeepCap~\cite{habermann2020deepcap} and DynaCap~\cite{habermann2021real} dataset sequences that are captured by dense camera views. MonoCap provides RGB images, foreground masks and camera parameters. We use EasyMocap to fit the SMPL parameters.

\section{Limitations and Discussions} \label{Limitations}
While our method offers a simple yet effective way of creating human avatars from very sparse view images, it still has a few limitations. First, it relies on accurate parametric body fitting with calibrated multi-view cameras. To democratize these human avatars, it will be worthwhile to explore directions which do not require explicit parametric model fitting. Second, the SMPL-driven deformation cannot express the complex non-rigid deformations (\eg, loose clothes or garments). Also, as the target pose deviates much from the available reference pose, our pose synthesis tends to degrade. It would be an fruitful direction to find an optimal reference frame (\eg, based on the similarity to a target pose). Other interesting future directions include scaling the approach to high resolution images and real-time rendering. Lastly, if the SMPL fit deviates much from the original geometry (in terms of pose or shape), the same SMPL vertice will be projected to non-corresponding pixels across different views/time. This will cause the blending of misaligned pixel colors, which in turn generates blurry results. Interesting future directions to improve the behavior are exploring the patch-wise perceptual loss as in GRAF~\cite{Schwarz2020NEURIPS} and Keypoint NeRF~\cite{mihajlovic2022keypointnerf} or to jointly optimize the SMPL fit. 

\section{Societal Impact} \label{Societal}
We discuss the potential societal impact of our work. The positive side of constructing an avatar from few number of cameras is that it can realize the immersive 3D teleconferencing that is affordable for most people, in contrast to earlier methods that require expensive reconstruction from dozens of cameras. Also, people can easily generate their own avatar, and explore the virtual world with it. The negative aspect is that it could be used in generating fake media, which can erode the trust against media. We strongly hope that our work can be used in positive directions.

\end{document}